%% file: CLiPPER.tex
\documentclass[11pt]{article}

\usepackage{acl2012}
\usepackage{times}
\usepackage{latexsym}
\usepackage{amsmath}
\usepackage{multirow}
\usepackage{url}
\usepackage{graphicx}
\usepackage{mathrsfs}
\usepackage{color}
\usepackage{CJKutf8}
\usepackage{bbm}
\usepackage{fmtcount}
\usepackage{color}
\usepackage{subfig}
\usepackage{tabularx}
\usepackage{float}
\usepackage{algorithmic,algorithm,eqparbox,array}
\usepackage{pgfplots}
\usepackage{filecontents}

\def\argmax{\underset{}{\operatorname{argmax}}}

\definecolor{gold}{rgb}{0.85,.66,0}
\definecolor{crimson}{RGB}{220,20,60}
\definecolor{indianred}{RGB}{176,23,31}
\definecolor{seagreen}{RGB}{67,205,128}

\def\bphi{\boldsymbol\phi}
\def\btheta{\boldsymbol\theta}
\def\by{\mathbf{y}}
\def\bx{\mathbf{x}}

\title{Cross-lingual Pseudo-Projected Expectation Regularization for \\  Weakly Supervised Learning}

\date{}

  \author{Mengqiu Wang \and Christopher D. Manning\\
 Computer Science Department \\
 Stanford University \\
   Stanford, CA 94305\ \ USA \\
 {\tt \{mengqiu,manning\}@cs.stanford.edu\ }}

\begin{document}
\maketitle
\begin{CJK}{UTF8}{gbsn}


\begin{abstract}
We consider a multilingual weakly supervised learning scenario where knowledge from annotated corpora in a resource-rich language is transferred via bitext to guide the learning in other languages. Past approaches project labels across bitext and use them as features or gold labels for training. We propose a new method that projects model expectations  rather than labels, which facilities transfer of model uncertainty across language boundaries. We encode expectations as constraints and train a discriminative CRF model using Generalized Expectation Criteria  \cite{Mann:2010:JMLR}. Evaluated  on standard Chinese-English and German-English NER datasets, our method demonstrates F$_1$ scores of 64\% and 60\% when no labeled data is used. Attaining the same accuracy with supervised CRFs requires 12k and 1.5k labeled sentences. Furthermore, when combined with labeled examples, our method yields significant improvements over state-of-the-art supervised methods, achieving best reported numbers to date on Chinese OntoNotes and German CoNLL-03 datasets.
\end{abstract}

\section{Introduction}
\label{sec:intro}

Supervised statistical learning methods have enjoyed great popularity in Natural Language Processing (NLP) over the past decade. 
The success of supervised methods depends heavily upon the availability of large amounts of annotated training data. 
Manual curation of  annotated corpora is a costly and time consuming process. To date, most annotated resources resides within the English language,  which hinders the adoption of  supervised learning methods in many multilingual environments.

To minimize the need for annotation, significant progress has been made in developing unsupervised and semi-supervised approaches to NLP (Collins and Singer 1999; Klein 2005; Liang 2005; Smith 2006; Goldberg 2010; \textit{inter alia})~\nocite{Collins:1999:EMNLP,Klein:2005:Thesis,Liang:2005:Thesis,Smith:2006:Thesis,Goldberg:2010:Thesis}.
More recent paradigms for semi-supervised learning allow modelers to directly encode knowledge about the task and the domain as constraints to guide learning \cite{Chang:2007:ACL,Mann:2010:JMLR,Ganchev:2010:JMLR}. 
However, in a multilingual setting, coming up with effective constraints require extensive knowledge of the foreign\footnote{For experimental purposes, we designate English as the resource-rich language, and other languages of interest as ``foreign''. In our experiments, we simulate the resource-poor scenario using Chinese and German, even though in reality these two languages are quite rich in resources.} language.

Bilingual parallel text (bitext)  lends itself as a medium to transfer knowledge from a resource-rich language to a foreign languages. 
\newcite{Yarowsky:2001:NAACL} project labels produced by an English tagger to the foreign side of bitext,  then use the projected labels to learn a HMM model. 
More recent work applied the projection-based approach to more language-pairs, and further improved performance through the use of type-level constraints from tag dictionary and feature-rich generative or discriminative models \cite{Das:2011:ACL,Tackstrom:2013:ACL}.  

In our work, we propose a new project-based method that differs in two important ways. 
First, we never explicitly project the labels. Instead, we project expectations over the labels. This pseudo-projection acts as a soft constraint over the labels,  which allows us to transfer more information and uncertainty across language boundaries.
Secondly, we encode the expectations as constraints and train a model by minimizing divergence between model expectations and projected expectations in a Generalized Expectation (GE) Criteria \cite{Mann:2010:JMLR} framework.

We evaluate our approach on Named Entity Recognition (NER) tasks for English-Chinese and English-German language pairs on standard public datasets. We report results in two settings: a weakly supervised setting where no labeled data or a small amount of labeled data is available, and a semi-supervised settings where labeled data is available, but we can gain predictive power by learning from unlabeled bitext.

\section{Related Work}\label{sec:related_work}

Most semi-supervised learning approaches embody the principle of learning from constraints. 
There are two broad categories of constraints: multi-view constraints, and external knowledge constraints.

Examples of methods that explore multi-view constraints include self-training \cite{Yarowsky:1995:ACL,McClosky:2006:NAACL},\footnote{A multi-view interpretation of self-training is that the self-tagged additional data offers new views to learners trained on existing labeled data.} co-training \cite{Blum:1998:COLT,Sindhwani:2005:ICML}, multi-view learning \cite{Ando:2005:ACL,Carlson:2010:WSDM}, and discriminative and generative model combination \cite{Suzuki:2008:ACL,Druck:2010:ICML}.

An early example of using knowledge as constraints in weakly-supervised learning is the work by \newcite{Collins:1999:EMNLP}. 
They showed that the addition of a small set of ``seed'' rules greatly improve a co-training style unsupervised tagger. 
\newcite{Chang:2007:ACL} proposed a constraint-driven learning (CODL) framework where constraints are used to guide the selection of best self-labeled examples to be included as additional training data in an iterative EM-style procedure. 
The kind of constraints used in applications such as NER are the ones like ``the words CA, Australia, NY are \textsc{Location}'' \cite{Chang:2007:ACL}. Notice the similarity of this particular constraint to the kinds of features one would expect to see in a discriminative model such as MaxEnt. The difference is that instead of learning the validity (or weight) of this feature from labeled examples --- since we do not have them --- we can constrain the model using our knowledge of the domain. 
\newcite{Druck:2009:EMNLP} also demonstrated that in an active learning setting where annotation budget is limited, it is more efficient to label features than examples. Other sources of knowledge include lexicons and gazetteers \cite{Druck:2007:NIPS,Chang:2007:ACL}.

While it is straight-forward to see how resources such as a list of city names can give a lot of mileage in recognizing locations, we are also exposed to the danger of over-committing to hard constraints. For example, it becomes problematic with city names that are ambiguous, such as Augusta, Georgia.\footnote{This is a city in the state of Georgia in USA, famous for its golf courses. It is ambiguous since both Augusta and Georgia can also be used as person names.} 
To soften these constraints, \newcite{Mann:2010:JMLR} proposed the Generalized Expectation (GE) Criteria framework, which encodes constraints as a regularization term over some score function that measures the divergence between the model's expectation and the target expectation.  
The connection between GE and CODL is analogous to the relationship between hard (Viterbi) EM and soft EM, as illustrated by \newcite{Samdani:2012:NAACL}.

Another closely related work is the Posterior Regularization (PR) framework by \newcite{Ganchev:2010:JMLR}. In fact, as \newcite{Bellare:2009:UAI} have shown, in a discriminative model these two methods optimize exactly the same objective.\footnote{The different terminology employed by GE and PR may be confusing to discerning readers, but the ``expectation'' in the context of GE means the same thing as ``marginal posterior'' as in PR.} The two differ in optimization details: PR uses a EM algorithm to approximate the gradients which avoids the expensive computation of a covariance matrix between features and constraints,  whereas GE directly calculates the gradient. However, later results \cite{Druck:2011:Thesis} have shown that using the Expectation Semiring techniques of \newcite{Li:2009:EMNLP}, one can compute the exact gradients of GE in a Conditional Random Fields (CRF) \cite{Lafferty:2001:ICML} at costs no greater than computing the  gradients of ordinary CRF. And empirically, GE tends to perform more accurately than PR \cite{Bellare:2009:UAI,Druck:2011:Thesis}.

Obtaining appropriate knowledge resources for constructing constraints remain as a bottleneck in applying GE and PR to new languages. 
However, a number of past work recognizes parallel bitext as a rich source of linguistic constraints, naturally captured in the translations. As a result, bitext has been effectively utilized for unsupervised multilingual grammar induction \cite{Alshawi:2000:MT,Snyder:2009:ACL}, parsing \cite{Burkett:2008:EMNLP}, and sequence labeling \cite{Naseem:2009:JMLR}.

A number of recent work also explored bilingual constraints in the context of simultaneous bilingual tagging, and showed that enforcing agreements between language pairs give superior results than monolingual tagging \cite{Burkett:2010:CONLL,Che:2013:NAACL,Wang:2013:AAAI}. They also demonstrated a \textit{uptraining} \cite{Petrov:2010:EMNLP} setting where tag-induced bitext can be used as additional monolingual training data to improve monolingual taggers. 
A major drawback of this approach is that it requires a readily-trained tagging models in each languages, which makes a weakly supervised setting infeasible. Another intricacy of this approach is  that it only works when the two models have comparable strength, since mutual agreements are enforced between them.

Projection-based methods can be very effective in weakly-supervised scenarios, as demonstrated by \newcite{Yarowsky:2001:NAACL}, and \newcite{Xi:2012:EMNLP}. One problem with  projected labels is that they are often too noisy to be directly used as training signals. To mitigate this problem, \newcite{Das:2011:ACL} designed a label propagation method to automatically induce a tag lexicon for the foreign language to smooth the projected labels. \newcite{Fossum:2005:IJCNLP} filter out projection noise by combining projections from from multiple source languages. However, this approach is not always viable since it relies on having parallel bitext from multiple source languages. \newcite{Li:2012:EMNLP} proposed the use of crowd-sourced Wiktionary as additional resources for inducing tag lexicons. More recently, \newcite{Tackstrom:2013:ACL} combined token-level and type-level constraints to constrain legitimate label sequences and and recalibrate the probability distribution in a CRF. The tag dictionary used for POS tagging are analogous to the gazetteers and name lexicons used for NER by \newcite{Chang:2007:ACL}.

Our work is also closely related to \newcite{Ganchev:2009:ICML}. They used a two-step projection method similar to \newcite{Das:2011:ACL} for dependency parsing.  Instead of using the projected linguistic structures as ground truth \cite{Yarowsky:2001:NAACL}, or as features in a generative model \cite{Das:2011:ACL}, they used them as constraints in a PR framework. 
Our work differs by projecting expectations rather than Viterbi one-best labels. We also choose the GE framework over PR. Experiments in \newcite{Bellare:2009:UAI} and \newcite{Druck:2011:Thesis} suggest that in a discriminative model (like ours), GE is more accurate than PR.

\begin{figure*}[t]
\centering
\includegraphics[width=.95\textwidth]{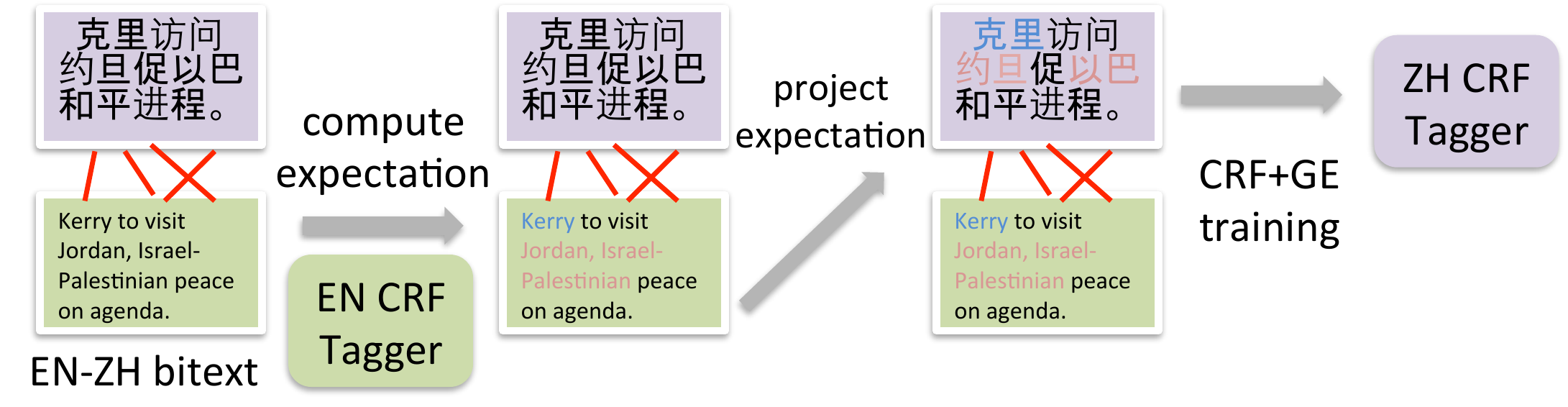}
\caption{Diagram illustrating the workflow of  {\bf C}ross-{\bf Li}ngual {\bf P}seudo-{\bf P}rojection {\bf E}xpectation {\bf R}egularization ({\bf CLiPPER}) method. Colors over the bitext are intended to denote model expectations, not the actual label assignments.}
 \label{fig:process}
\end{figure*}

\section{Approach}
Given bitext between English and a foreign language, our goal is to learn a CRF model in the foreign language from little or no labeled data. 
Our method performs {\bf C}ross-{\bf Li}ngual {\bf P}seudo-{\bf P}rojection {\bf E}xpectation {\bf R}egularization ({\bf CLiPPER}).

Figure~\ref{fig:process} illustrates the high-level workflow.
For every aligned sentence pair in the bitext, we first compute the posterior marginal at each word position on the English side using a pre-trained English CRF tagger; then for each aligned English word, we project its posterior marginal as expectations to the aligned word position on the foreign side.
 
We would like to learn a CRF model in the foreign language that has similar expectations as the projected expectations from English. 
To this end, we adopt the Generalized Expectation (GE) Criteria framework introduced by \newcite{Mann:2010:JMLR}.
In the remainder of this section, we follow the notation used in \cite{Druck:2011:Thesis} to explain our approach.

\subsection{CLiPPER}
The general idea of GE is that we can express our preferences over models through constraint functions. 
A desired model should satisfy the imposed constraints by matching the expectations on these constraint functions with some target expectations (attained by external knowledge like lexicons or in our case transferred knowledge from English). 
We define a constraint function $\bphi_{i, l_j}$  for each word position $i$ and output label assignment $l_j$ as a label identity indicator: 
$$
\phi_{i, l_j}(\by) = \left\{ \begin{array}{rl}
 1 &\mbox{ if $l_j = y_i$ and $A_i \neq \emptyset$} \\
  0 &\mbox{ otherwise}
       \end{array} \right.
$$
The set $\{l_1, \cdots, l_m\}$ denotes all possible label assignment for each $y_i$, and $m$ is number of label values.
$A_i$ is the set of English words aligned to Chinese word $i$. 
The condition $A_i \neq \emptyset$ specifies that the constraint function applies only to Chinese word positions that have at least one aligned English word. 
Each $\phi_{i, l_j}(\by)$ can be treated as a Bernoulli random variable, and we concatenate the set of all $\phi_{i, l_j}$ into a  \textit{random vector} $\bphi(\by)$, where $\bphi_k = \phi_{i, l_j}$ if $k = i * m + j$.
We drop the $(\by)$ in $\bphi$ for simplicity.

The target expectation over $\phi_{i, l_j}$, denoted as $\tilde\phi_{i, l_j}$, is the expectation of assigning label $l_j$ to English word $A_i$\footnote{An English word aligned to foreign word at position $i$. When multiple English words are aligned to the same foreign word, we average the expectations.} under the English conditional probability model.

The expectation over $\bphi$ under a conditional probability model $P(\by|\bx;{\btheta})$ is denoted as $E_{P(\by|\bx;{\btheta})}[\bphi]$, and simplified as $E_{\btheta}[\bphi]$ whenever it is unambiguous.

The conditional probability model $P(\by|\bx;{\btheta})$ in our case is defined as a standard linear-chain CRF:\footnote{We simplify notation by dropping the $L_2$ regularizer in the CRF definition, but apply it in our experiments.}
\begin{align*}
	P(\by|\bx;{\btheta}) = \frac{1}{Z(\bx;\btheta)}{\exp\left( \sum\limits_{i}^{n}{\btheta \mathbf{f}({\bx, y_i, y_{i-1}}) }  \right)}
\end{align*} 
\noindent where $\mathbf{f}$ is a set of feature functions; $\btheta$ are the matching parameters to learn; $n = |\bx|$.

The objective function to maximize in a standard CRF is the log probability over a collection of labeled documents:
\begin{align}
	L_{CRF}(\btheta) = \sum\limits_{a=1}^{a'}\log P(\by_{a}^{*}|\bx_{a};{\btheta})
 \label{eqn:obj_crf}
\end{align} 
\noindent $a'$ is the number of labeled sentences. $\by^*$ is an observed label sequence.

The objective function to maximize in GE is defined as the sum over all unlabeled examples (foreign side of bitext), over some cost function $S$ between between the model expectation ($E_{\btheta}[\bphi] $) and the target expectation ($\tilde\bphi$) over $\bphi$. 

We choose $S$ to be the negative $L_2^2$  squared error,\footnote{In general, other loss functions such as  KL-divergence can also be used for $S$. We found $L_2^2$ to work well in practice.} defined as:
\begin{align}
	L_{GE}(\btheta) &= \sum\limits_{b=1}^{n'} S\left(E_{P(\by|\bx_b;\btheta)}[\bphi(\by_b)], \tilde\bphi(\by_b\right) \nonumber \\
	& = \sum\limits_{b=1}^{b'} -\| \tilde\bphi(\by_b) - E_{\btheta}[\bphi(\by_b)]  \|_2^2 
\label{eqn:obj_ge}
\end{align}
\noindent $b'$ is the total number of unlabeled bitext sentence pairs. 

When both labeled and bitext training data are available, the joint objective is the sum of Eqn.~\ref{eqn:obj_crf} and \ref{eqn:obj_ge}. Each is computed  over the labeled training data and foreign half in the bitext, respectively.

We can optimize this joint objective by computing the gradients and use a gradient-based optimization method such as L-BFGS.
Gradients of $L_{CRF}$ decomposes down to the gradients over each labeled training example $(\bx, \by^{*})$, computed as:
\begin{align*}
\vspace{-3cm}
\frac{\partial}{\partial \btheta}(\log P(\by^{*}_{a}|\bx_{a};{\btheta}) = \tilde{E}[\btheta] - E[\btheta] \\
\vspace{-3cm}
\end{align*}
\noindent where  $\tilde E[\btheta]$ and $E[\btheta]$ are the empirical and expected feature counts, respectively.

Computing the gradient of $L_{GE}$ decomposes down to the gradients of $S(E_{P(\by|\bx_b;\btheta}[\bphi])$ for each unlabeled foreign sentence $\bx$ and the constraints over this example $\bphi$ . The gradients can be calculated as:
\begin{align*}
\frac{\partial}{\partial \btheta}S(E_{\btheta}[\bphi] ) & =  -\frac{\partial}{\partial \btheta}\left( \tilde\bphi -  E_{\btheta}[\bphi] \right)^{T} \left(  \tilde\bphi -  E_{\btheta}[\bphi] \right) \\
& = 2\left( \tilde\bphi -  E_{\btheta}[\bphi] \right)^{T} \left( \frac{\partial}{\partial \btheta} E_{\btheta}[\bphi] \right)
\end{align*}
We redefine the penalty vector $\mathbf{u} = 2\left( \tilde\bphi -  E_{\btheta}[\bphi] \right)$ to be $u$.
$\frac{\partial}{\partial \btheta} E_{\btheta}[\bphi]$ is a matrix where each column contains the gradients for a particular model feature $\theta$ with respect to all constraint functions $\bphi$. It can be computed as:
\begin{align*}
&\frac{\partial}{\partial \btheta} E_{\btheta}[\bphi] =  \sum\limits_{\by} \bphi(\by) \frac{\partial}{\partial \btheta} P(\by|\bx;\btheta)\nonumber \\
=&  \sum\limits_{\by} \bphi(\by) \frac{\partial}{\partial \btheta} \left(  \frac{1}{Z(\bx;\btheta)}  {\exp({\btheta^{T}\mathbf{f}({\bx,\by}))}} \right) \nonumber \\
=&   \sum\limits_{\by} \bphi(\by) \Bigg( \frac{1}{Z(\bx;\btheta)} \left(  \frac{\partial}{\partial\btheta} {\exp({\btheta^{T}\mathbf{f}({\bx,\by}))}}  \right) \nonumber \\
& + {\exp({\btheta^{T}\mathbf{f}({\bx,\by}))}} \left( \frac{\partial}{\partial\btheta} \frac{1}{Z(\bx;\btheta)}  \right) \Bigg) \nonumber
\end{align*}
\begin{align}
= &  \sum\limits_{\by} \bphi(\by) \bigg( P(\by | \bx; \btheta) \mathbf{f}({\bx,\by})^{T}  \nonumber\\
& -  P(\by | \bx; \btheta)\sum_{\by'}  P(\by' | \bx; \btheta)  \mathbf{f}({\bx,\by'})^{T} \bigg) \nonumber \\
=&   \sum\limits_{\by} P(\by | \bx; \btheta)\sum_{\by}  \bphi(\by)  \mathbf{f}({\bx,\by})^{T} \nonumber \\
&-  \big( \sum\limits_{\by} P(\by | \bx; \btheta) \bphi(\by)  \big) \big( \sum_{\by}  P(\by | \bx; \btheta)  \mathbf{f}({\bx,\by})^{T} \big) \nonumber \\
 = & \; \; \mathrm{COV}_{P(\by|\bx;\btheta)}\left( \bphi(\by), \mathbf{f}(\bx,\by) \right) \label{eqn:cov}\\
=  & \; \; E_{\btheta}[\bphi\mathbf{f}^{T}] - E_{\btheta}[\bphi]E_{\btheta}[\mathbf{f}^{T}] \label{eqn:before}
\end{align}
Eqn.~\ref{eqn:cov} gives the intuition of how optimization works in GE. In each iteration of L-BFGS, the model parameters are updated according to their covariance with the constraint features, scaled by the difference between current expectation and target expectation. 
The term $E_{\btheta}[\bphi\mathbf{f}^{T}]$ in Eqn.~\ref{eqn:before} can be computed using a dynamic programming (DP) algorithm, but solving it directly requires us to store a matrix of the same dimension as $\mathbf{f}^{T}$ in each step of the DP. 
We can reduce the complexity by using the following trick:
\begin{align}
& \frac{\partial}{\partial \btheta}S(E_{\btheta}[\bphi] ) =  u^{T} \left( \frac{\partial}{\partial \btheta} E_{\btheta}[\bphi] \right) \nonumber \\
= &  \mathbf{u}^{T} \left( E_{\btheta}[\bphi\mathbf{f}^{T}] - E_{\btheta}[\bphi]E_{\btheta}[\mathbf{f}^{T}] \right) \nonumber \\
= &  E_{\btheta}[\mathbf{u}^{T} \bphi\mathbf{f}^{T}] - E_{\btheta}[\mathbf{u}^{T} \bphi]E_{\btheta}[\mathbf{f}^{T}] \nonumber \\
= &  E_{\btheta}[\bphi'\mathbf{f}^{T}] - E_{\btheta}[\bphi']E_{\btheta}[\mathbf{f}^{T}] \label{eqn:after} \\
& \bphi' = \mathbf{u}^{T} \bphi \nonumber
\end{align}
Now in Eqn.~\ref{eqn:after}, $E_{\btheta}[\bphi']$ becomes a scalar value; and to compute the term $E_{\btheta}[\bphi'\mathbf{f}^{T}] $, we only need to store a vector in each step of the following DP algorithm  \cite[93]{Druck:2011:Thesis}: 
\begin{align*}
E_{\btheta}[\bphi'\mathbf{f}^{T}]   = & \sum\limits_{i=1}^{n}\sum\limits_{y_i}\sum\limits_{y_{i+1}}\bigg\{ \Big[ \sum\limits_{j=1}^{n} \sum\limits_{y_j} P(y_i, y_{i+1}, y_j | \bx) \\
\cdot & \;  \bphi(y_j) \Big] \cdot  \mathbf{f}(y_i, y_{i+1}, \bx)^{T} \bigg\}
\end{align*}
The bracketed term can be broken down to two parts:
\begin{align*}
 &\sum\limits_{j=1}^{n} \sum\limits_{y_j} P(y_i, y_{i+1}, y_j | \bx) \bphi(y_j)  \\
= &\sum\limits_{j=1}^{i} \sum\limits_{y_j} P(y_i, y_{i+1}, y_j | \bx) \bphi(y_j) \\
+ & \sum\limits_{j=i+1}^{n} \sum\limits_{y_j} P(y_i, y_{i+1}, y_j | \bx) \bphi(y_j) \\
= & \; \alpha(y_i, y_{i+1}, i) + \beta(y_i, y_{i+1}, i)
\end{align*}
\begin{align*}
& \alpha(y_0, y_1, 0) \equiv  P(y_0, y_1 | \bx) \bphi(y_0) \\
& \alpha(y_i, y_{i+1}, i) \equiv  P(y_i, y_{i+1} | \bx) \bphi(y_i) + \\
& P(y_{i+1} | y_i, \bx) \sum\limits_{y_{i-1}} \alpha(y_{i-1}, y_i, i-1) \\
& \beta(y_{n-1}, y_n, n-1) \equiv P(y_{n-1}, y_n | \bx) \bphi(y_n) \\
& \beta(y_{i}, y_{i+1}, i) \equiv P(y_{i}, y_{i+1} | \bx) \bphi(y_{i+1}) + \\
& P(y_{i} | y_{i+1}, \bx) \sum\limits_{y_{i+2}} \beta(y_{i+1}, y_{i+2}, i+1)
\end{align*}

The resulting algorithm has complexity $O(nm^2)$, which is the same as the standard forward-backward inference algorithm for CRF.

\subsection{Hard vs. Soft Projection}\label{sec:hard}
Projecting expectations instead of  one-best label assignments from English to foreign language can be thought of as a soft version of the method described in \cite{Das:2011:ACL} and \cite{Ganchev:2009:ICML}. Soft projection has its advantage: when the English model is not certain about its predictions,  we do not have to commit to the current best prediction. The foreign model has more freedom to form its own belief since any marginal distribution it produces would deviates from a flat distribution by just about the same amount. In general, preserving uncertainties till later is a strategy that  has benefited many NLP tasks \cite{Finkel:2006:EMNLP}.
Hard projection can also be treated as a special case in our framework. We can simply recalibrate posterior marginal of English by assigning probability mass $1$ to the most likely outcome, and zero everything else out, effectively taking the $\argmax$ of the marginal at each word position. We refer to this version of expectation as the ``hard'' expectation.
In the hard projection setting, GE training resembles a ``project-then-train'' style semi-supervised CRF training scheme \cite{Yarowsky:2001:NAACL,Tackstrom:2013:ACL}. In such a training scheme, we project the one-best predictions of English CRF to the foreign side through word alignments, then include the newly ``tagged'' foreign data as additional training data to a standard CRF in the foreign language.
The difference between GE training and this scheme  is that they optimize different objectives: 
CRF optimizes maximum conditional likelihood of the observed label sequence, whereas GE minimizes squared error between model's expectation and ``hard'' expectation based on the observed label sequence. 
We compare the hard and soft variants of GE with the project-then-train style CRF training in our experiments and report results in Section~\ref{sec:min-results}. 

\section{Experiments}
We conduct experiments on Chinese and German NER. We evaluate CLiPPER in two learning settings: weakly supervised  and semi-supervised. In the weakly supervised setting, we simulate the condition of having no labeled training data, and evaluate the model learned from bitext alone. We then vary the amount of labeled data available to the model, and examine the model's learning curve. In the semi-supervised setting, we assume our model has access to the full labeled data; our goal is to improve performance of the supervised method by learning from additional bitext.

\subsection{Dataset and Setup}
We used the latest version of Stanford NER Toolkit\footnote{{\url{http://www-nlp.stanford.edu/ner}}} as our base CRF model in all experiments. Features for English, Chinese and German CRFs are documented extensively in \cite{Che:2013:NAACL} and \cite{Faruqui:2010:KONVENS} and omitted here for brevity. 
It it worth noting that the  current Stanford NER models include recent improvements from semi-supervise learning approaches that induces distributional similarity features from large word clusters. These models represent the current state-of-the-art in supervised methods, and serve as a very strong baseline. 

For Chinese NER experiments, we follow the same  setup as \newcite{Che:2013:NAACL} to evaluate on the latest  OntoNotes (v4.0) corpus \cite{Hovy:2006:NAACL}.\footnote{LDC catalogue No.: LDC2011T03} 
A total of 8,249 sentences from the parallel Chinese and English Penn Treebank portion \footnote{File numbers: chtb\_0001-0325, ectb\_1001-1078} are reserved for evaluation. 
Odd-numbered documents are used as development set, and even-numbered documents are held out as blind test set. 
The rest of OntoNotes annotated with NER tags are used to train the English and Chinese CRF base taggers.
There are about 16k and 39k labeled sentences for Chinese and English training, respectively. 
The English CRF tagger trained on this training corpus gives F$_1$ score of 81.68\% on the OntoNotes test set.
Four entities types (\textsc{Person}, \textsc{Location}, \textsc{Organization} and \textsc{GPE}) are used with a BO tagging scheme.
The English-Chinese bitext comes from the Foreign Broadcast Information Service corpus (FBIS).\footnote{LDC catalogue No.: LDC2003E14}
 It is first sentence aligned using the Champollion Tool Kit,\footnote{{\url{champollion.sourceforge.net}}} then word aligned with the BerkeleyAligner.\footnote{{\url{code.google.com/p/berkeleyaligner}}}

For German NER experiments, we evaluate using the standard CoNLL-03 NER corpus \cite{Sang:2003:CoNLL}. The labeled training set has 12k and 15k sentences. We used the de-en portion of the \textit{News Commentary}\footnote{{\url{http://www.statmt.org/wmt13/training-parallel-nc-v8.tgz}}} data from WMT13 as bitext. 
The English CRF tagger trained on CoNLL-03 English training corpus gives F$_1$ score of 90.4\% on the CoNLL-03 test set.

We report standard entity-level precision (P), recall (R) and F$_1$ score given by \textsc{ConllEval} script on both the development and test sets. Statistical significance tests are done using a paired bootstrap resampling method with 1000 iterations, averaged over 5 runs.
We compare against three recently approaches that were introduced in Section~\ref{sec:related_work}. They are: semi-supervised learning method using factored bilingual models with Gibbs sampling \cite{Wang:2013:AAAI}; 
bilingual NER using Integer Linear Programming (ILP) with bilingual constraints, by \cite{Che:2013:NAACL}; and constraint-driven bilingual-reranking approach \cite{Burkett:2010:CONLL}.
The code from \cite{Che:2013:NAACL} and \cite{Wang:2013:AAAI} are publicly available,\footnote{{\url{https://github.com/stanfordnlp/CoreNLP}}}. Code from \cite{Burkett:2010:CONLL} is obtained through personal communications.\footnote{Due to technical difficulties, we are unable to replicate \newcite{Burkett:2010:CONLL} experiments on German NER, therefore only Chinese results are reported.}

Since the objective function in Eqn.~\ref{eqn:obj_ge} is non-convex, we adopted the early stopping training scheme from \cite{Turian:2010:ACL} as the following: after each iteration in L-BFGS training, the model is evaluated against the development set; the training procedure is terminated if no improvements have been made in 20 iterations. 

\subsection{Weakly Supervised Results}
\label{sec:min-results}
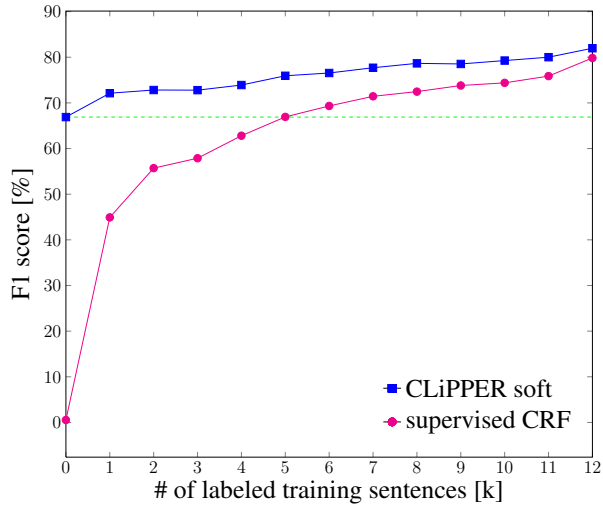
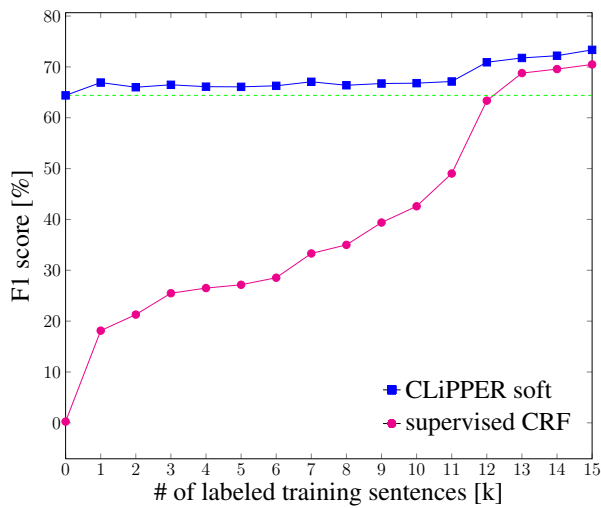
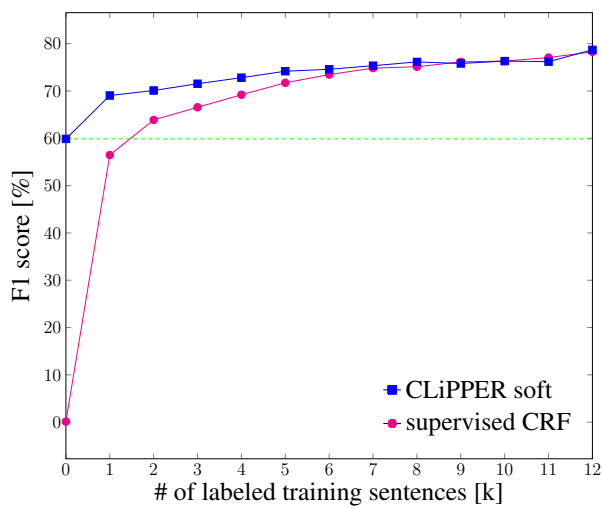
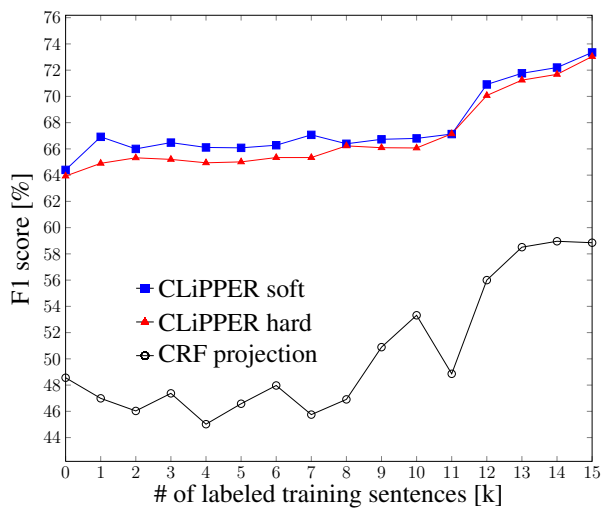
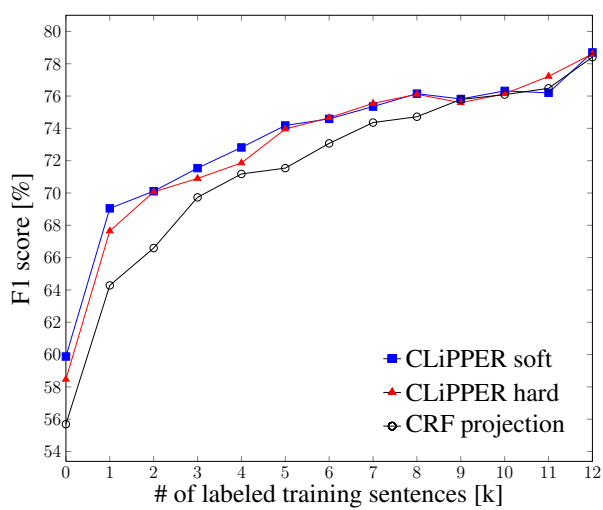
\begin{figure*}
\centering
\subfloat[Chinese Dev]{
\resizebox{\columnwidth}{!} {\input{cn_dev}} 
\label{fig:curve_cn_dev}}
\subfloat[German Dev]{
\resizebox{\columnwidth}{!} {\input{de_dev}} 
\label{fig:curve_de_dev}}

\subfloat[Chinese Test]{
\resizebox{\columnwidth}{!} {\input{cn_test}} 
\label{fig:curve_cn_test}}
\subfloat[German Test]{
\resizebox{\columnwidth}{!} {\input{de_test}} 
\label{fig:curve_de_test}}

\subfloat[Soft vs. Hard on Chinese Test]{
\resizebox{\columnwidth}{!} {\input{cn_test_sVSh}} 
\label{fig:curve_cn_dev_sVSh}}
\subfloat[Soft vs. Hard on German Test]{
\resizebox{\columnwidth}{!} {\input{de_test_sVSh}} 
\label{fig:curve_de_dev_sVSh}}
\caption{Performance curves of CLiPPER with varying amounts of available labeled training data in a weakly supervised setting. Vertical axes show the F$_1$ score on the development and test set, respectively. Performance curves of supervised CRF and ``project-then-train'' CRF are plotted for comparison.}
\label{fig:learning_curve}
\end{figure*}

The top four figures in Figure~\ref{fig:learning_curve} show results of weakly supervised learning experiments. Quite remarkably, on Chinese test set, our proposed method (CLiPPER) achieves a F$_1$ score of 64.4\% with 80k bitext, when no labeled training data is used. In contrast, the supervised CRF baseline would require as much as 12k labeled sentences to attain the same accuracy. 
Results on the German test set is less striking. With no labeled data and 40k of bitext, CLiPPER performs at F$_1$ of 60.0\%, the equivalent of using 1.5k labeled examples in the supervised setting. When combined with 1k labeled examples, performance of CLiPPER reaches 69\%,  a gain of over 5\% absolute over supervised CRF.
We also notice that supervised CRF model learns much faster in German than Chinese. This result is not too surprising, since it is well recognized that Chinese NER is more challenging than German or English due to the lack of orthographical features, such as word capitalization. Chinese NER relies more on lexicalized features, and therefore needs more labeled data to achieve good coverage. The results also suggest that CLiPPER seems to be very effective at transferring lexical knowledge from English to Chinese.

The bottom two figures in Figure~\ref{fig:learning_curve} compares soft GE projection with hard GE projection and the ``project-then-train'' style CRF training scheme (\textit{cf.} Section~\ref{sec:hard}). We observe that both soft and hard GE projection significantly outperform the ``project-then-train'' style training scheme. The difference is especially pronounced on the Chinese results when fewer labeled examples are available. Soft projection gives better accuracy than hard projection when no labeled data is available, and also has a faster learning rate.

\subsection{Semi-supervised Results}
\begin{table}[t]
\setlength{\tabcolsep}{4pt}
\begin{scriptsize}
\begin{center}
\renewcommand{\arraystretch}{1.1}
\begin{tabular}{ r | c | c | c | c | c | c | c }
  \multicolumn{2}{c|}{}  & \multicolumn{3}{c|}{{\normalsize Chinese}} & \multicolumn{3}{c}{{\normalsize German}} \\
 \multicolumn{1}{c}{} &  & P & R & F$_1$ & P & R & F$_1$ \\
\hline
\multicolumn{2}{c|}{CRF}                   &  79.87 & 63.62  & 70.83  & 88.05 & 73.03 &  79.84 \\
\hline
\multirow{4}{*}{CLiPPER$_s$} & 10k      &   81.36 & 65.16 & 72.36    &  85.23  &77.79 & 81.34  \\
                                       & 20k       &  \bf{81.79}  & 64.80 &  72.31    & 88.11  & 75.93  & 81.57 \\
                                       &  40k      &  79.24  & \bf{66.08}  & 72.06   &  \bf{88.25} & 76.52 & \bf{81.97} \\
                                       &  80k      &  80.26  & 65.92  & \bf{72.38}   &  87.80 & \bf{76.82} & 81.94
\end{tabular}
\end{center}
\caption{Chinese and German NER results  on the development set using CLiPPER with varying amounts of unlabeled bitext (10k, 20k, etc.). Best number of each column is highlighted in bold. 
The F$_1$ score improvements over CRF baseline in all cases are statistically significant at 99.9\% confidence level. }
\label{tbl:dev_results}
\end{scriptsize}
\end{table}

In the semi-supervised experiments, we let the CRF model use the full set of labeled examples in addition to the unlabeled bitext. 
Table~\ref{tbl:dev_results} shows results on the development dataset for Chinese and German using 10-80k bitext.
We see that with merely 10k additional bitext, CLiPPER is able to improve significantly over state-of-the-art CRF baselines by as much as 1.5\% F$_1$ on both Chinese and German. 
With more unlabeled data, we notice a tradeoff between precision and recall on Chinese. The final F$_1$ score on Chinese at 80k level is only marginally better than 10k. On the other hand, we observe a modest but steady improvement on German as we add more unlabeled bitext, up until 40k sentences. 
We select the best configurations on development set (80k for Chinese and 40k for German) to evaluate on test set.

\begin{table}[t]
\setlength{\tabcolsep}{4pt}
\begin{scriptsize}
\begin{center}
\renewcommand{\arraystretch}{1.1}
\begin{tabular}{ l  | l | l | l | l | l | l }
  \multicolumn{1}{c|}{}  & \multicolumn{3}{c|}{{\normalsize Chinese}} & \multicolumn{3}{c}{{\normalsize German}} \\
  &  \multicolumn{1}{c|}{P} &  \multicolumn{1}{c|}{R} &  \multicolumn{1}{c|}{F$_1$} &  \multicolumn{1}{c|}{P} &  \multicolumn{1}{c|}{R} &  \multicolumn{1}{c}{F$_1$} \\
\hline
CRF                   &  79.09 & 63.56 & 70.48  & 86.77  & 71.30 & 78.28 \\ 
CRF$_{ptt}$       &  {\bf 84.01} & 45.29 &  58.85  & 81.50 & {\bf 75.56} &  78.41 \\ 
\hline
WCD13            &  {80.31} & {65.78} & {72.33}  & {85.98} &   {72.37} & {78.59} \\       
CWD13            &  81.31  &  {65.50} & {72.55}  &   {85.99} &   { 72.98} &  {78.95}  \\ 
BPBK10           & 79.25 & 65.67 &  71.83 &  - & - & - \\
\hline
CLiPPER$_{h}$     & {83.67} & 64.80 &  73.04$^{\S\ddag}$ &  86.52 & 72.02 & 78.61 \\
CLiPPER$_{s}$     &  82.57 & \bf{65.99} & \bf{73.35}$_{\diamond\ast}^{\S\dag\star}$  &   \bf{87.11} & 72.56 &  \bf{79.17}$^{\S\ddag\star}$

\end{tabular}
\end{center}
\caption{Chinese and German NER results  on the test set. Best number of each column is highlighted in bold. \textsc{CRF} is the supervised baseline. \textsc{CRF$_{ptt}$} is the ``project-then-train'' semi-supervised scheme for CRF.
\textsc{WCD13} is \cite{Wang:2013:AAAI}, \textsc{CWD13} is \cite{Che:2013:NAACL}, and  \textsc{BPBK10} is \cite{Burkett:2010:CONLL}.  \textsc{CLiPPER$_s$} and \textsc{CLiPPER$_h$} are the soft and hard projections.
 $\S$ indicates F$_1$ scores that are statistically significantly better than CRF baseline at 99.5\% confidence level; $\star$ marks significance over \textsc{CRF$_{ptt}$} with 99.5\% confidence; $\dagger$ and $\ddagger$ marks significance over \textsc{WCD13} with 99.9\% and 94\% confidence; and $\diamond$ marks significance over \textsc{CWD13} with 99.7\% confidence; $\ast$ marks significance over \textsc{BPBK10} with 99.9\% confidence. }
\label{tbl:test_results}
\end{scriptsize}
\end{table}

\begin{figure*}
\centering
\subfloat[][]{
\includegraphics[width=.7\textwidth]{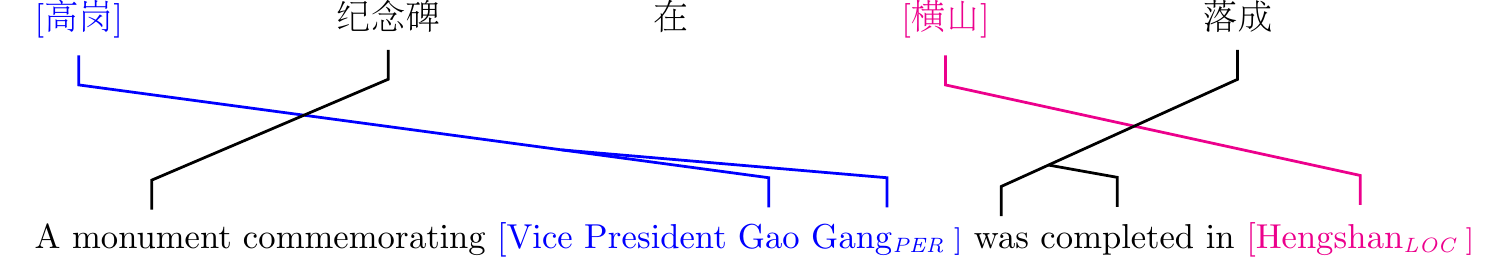}
 \label{fig:example_PER}}
 
\subfloat[Example where word proceeding ``monument'' is of type LOCATION][]{
\includegraphics[width=.7\textwidth]{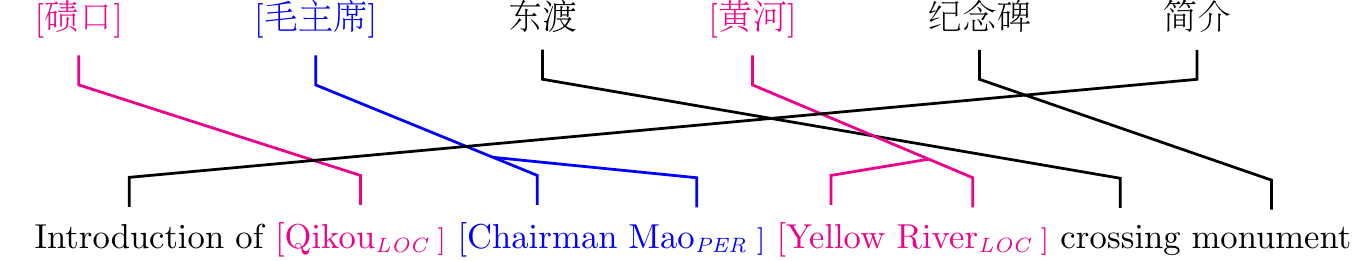}
\label{fig:example_LOC}}
\caption{Examples of aligned sentence pairs in Chinese and English. The lines going across a sentence pair indicate individual word alignments induced by an automatic word aligner. Entities of type \textsc{Location} are highlighted in \textit{\textcolor{magenta}{magenta}}, and entities of type \textsc{Person} are highlighted in \textit{\textcolor{blue}{blue}}.}
\label{fig:example}
\end{figure*}

Results on the test set are shown in Table~\ref{tbl:test_results}. 
All semi-supervised baselines are tested with the same number of unlabeled bitext as CLiPPER in each language.
The ``project-then-train'' semi-supervised training scheme severely hurts performance on Chinese, but gives a small improvement on German. Moreover, on Chinese it learns to achieve high precision but at a significant loss in recall. On German its behavior is the opposite. Such drastic and erratic imbalance suggest that this method is not robust or reliable.
The other three semi-supervised baselines (row 3-5) all show improvements over the CRF baseline, consistent with their reported results. 
\textsc{CLiPPER$_s$} gives the best results on both Chinese and German, yielding statistically significant improvements over all baselines except for \textsc{CWD13} on German. 
The hard projection version of CLiPPER also gives sizable gain over CRF. However,  in comparison, \textsc{CLiPPER$_s$} is superior.

The improvements of \textsc{CLiPPER$_s$} over CRF on Chinese test set is over 2.8\% in absolute F$_1$. The improvement over CRF on German is almost a percent. To our knowledge, these are the best reported numbers on the OntoNotes Chinese and CoNLL-03 German datasets.

\subsection{Efficiency}
Another advantage of our proposed approach is efficiency. Because we eliminated the previous multi-stage ``project-then-train'' paradigm, but instead integrating the semi-supervised and supervised objective into one joint objective, we are able to attain significant speed improvements. Table~\ref{tbl:time} shows the training time required to produce models that give results in Table~\ref{tbl:test_results}. 

\begin{table}
\setlength{\tabcolsep}{5pt}
\begin{small}
\begin{center}
\renewcommand{\arraystretch}{1.1}
\begin{tabular}{ l  | l | l }
              & \multicolumn{1}{c|}{Chinese}   & \multicolumn{1}{c}{German}\\
              \hline
CRF                   &  19m30s  & 7m15s   \\ 
CRF$_{ptt}$       &  34m2s & 12m45s \\ 
\hline
WCD13            & 3h17m   & 1h1m  \\       
CWD13            &  16h42m & 4h49m \\ 
BPBK10           & 6h16m & -  \\
\hline
CLiPPER$_{h}$     &     1h28m   &  16m30s \\
CLiPPER$_{s}$     &   1h40m & 18m51s
\end{tabular}
\end{center}
\caption{Timing stats during model training. } 
\label{tbl:time}
\end{small}
\end{table}

\section{Error Analysis and Discussion}
Figure~\ref{fig:example} gives two examples of CLiPPER in action. Both examples have a named entity that immediately proceeds the word ``纪念碑'' (monument) in the Chinese sentence. In Figure~\ref{fig:example_PER}, the word ``高岗'' has literal meaning of \textit{a hillock located at a high position}, which also happens to be the name of a former vice president of China. Without having previously observed this word as a person name in the labeled training data, the CRF model does not have enough evidence to believe that this is a \textsc{Person},  instead of \textsc{Location}. But the aligned words in English (``Gao Gang'') are clearly part of a person name as they were preceded by a title (``Vice President''). The English model has high expectation that the aligned Chinese word of "Gao Gang" is also a \textsc{Person}. Therefore,  projecting the English expectations to Chinese provides a strong clue to help disambiguating this word. Figure~\ref{fig:example_LOC} gives another example:  the word ``黄河''(Huang He, the Yellow River of China) can be confused with a person name since ``黄''(Huang or Hwang) is also a common Chinese last name.\footnote{In fact, a people search of the name 黄河 on the Chinese equivalent of Facebook (www.renren.com) returns over 13,000 matches.}. Again, knowing the translation in English, which has the indicative word ``River'' in it, helps disambiguation. 

\section{Conclusion}\label{sec:conc}

We introduced a domain and language independent semi-supervised method for training discriminative models by projecting expectations across bitext. Experiments on Chinese and German NER show that our method, learned over bitext alone, can rival performance of supervised models trained with thousands of labeled examples. Furthermore, applying our method in a setting where all labeled examples are available also shows improvements over state-of-the-art supervised methods. Our experiments also showed that soft expectation projection is more favorable to hard projection. This technique can be generalized to all sequence labeling tasks, and can be extended to include more complex constraints. For future work, we plan to apply this method to more language pairs and examine the formal properties of the model. 

%

\bibliographystyle{acl2012}
\bibliography{CLiPPER}
\end{CJK}
\end{document}

%% file: cn_dev.tex

\begin{tikzpicture}
\pgfplotsset{
  tick label style = {font=\Large},
  every axis label = {font=\Large},
  legend style = {font=\huge},
  label style = {font=\huge}
}

\begin{axis}[ width=\linewidth, ytick scale label code/.code={},
xmin=0,xmax=15,
        xtick=data, xlabel={\# of labeled training sentences [k]}, 
                ylabel={F1 score [\%]}]
\addplot[mark=*, mark options={fill=white,magenta},style={magenta, /tikz/mark size=3}] table [x=S, y=F1] {cn_crf.data};
\addplot[mark=square*, mark options={fill=black,blue}, style={blue, /tikz/mark size=3}] table [x=S, y=F1] {cn_ge.data};
\addplot [green, no markers, dashed] coordinates {(-1,63.44) (16,63.44)};
\end{axis}

\begin{scope}[shift={(9,1)}] 
\draw (0,0) -- 
  plot[mark=*, mark options={fill=white,magenta}, style={magenta,/tikz/mark size=3}] (0.25,0) -- (0.5,0) 
  node[right,font=\huge]{supervised CRF};
\draw[yshift=\baselineskip*2] (0,0) -- 
  plot[mark=square*, mark options={fill=black,blue},style={blue,/tikz/mark size=3}] (0.25,0) -- (0.5,0)
  node[right, font=\huge]{CLiPPER soft};
\end{scope}

\end{tikzpicture}

%% file: de_dev.tex

\begin{tikzpicture}
\pgfplotsset{
  tick label style = {font=\Large},
  every axis label = {font=\Large},
  legend style = {font=\huge},
  label style = {font=\huge}
}

\begin{axis}[ width=\linewidth, ytick scale label code/.code={},
  xmin=0, xmax=12,
        xtick=data, xlabel={\# of labeled training sentences [k]}, 
                ylabel={F1 score [\%]}]
\addplot[mark=*, mark options={fill=white,magenta},style={magenta,/tikz/mark size=3}] table [x=S, y=F1] {de_crf.data};
\addplot[mark=square*, mark options={fill=black,blue}, style={blue,/tikz/mark size=3}] table [x=S, y=F1] {de_ge.data};
\addplot [green, no markers, dashed] coordinates {(-1,66.89) (13,66.89)};
\end{axis}

\begin{scope}[shift={(9,1)}] 
\draw (0,0) -- 
  plot[mark=*, mark options={fill=white,magenta}, style={magenta,/tikz/mark size=3}] (0.25,0) -- (0.5,0) 
  node[right,font=\huge]{supervised CRF};
\draw[yshift=\baselineskip*2] (0,0) -- 
  plot[mark=square*, mark options={fill=black,blue},style={blue,/tikz/mark size=3}] (0.25,0) -- (0.5,0)
  node[right, font=\huge]{CLiPPER soft};
\end{scope}

\end{tikzpicture}

%% file: cn_test.tex

\begin{tikzpicture}
\pgfplotsset{
  tick label style = {font=\Large},
  every axis label = {font=\Large},
  legend style = {font=\huge},
  label style = {font=\huge}
}

\begin{axis}[ width=\linewidth, ytick scale label code/.code={},
    xmin=0, xmax=15,
        xtick=data, xlabel={\# of labeled training sentences [k]}, 
                ylabel={F1 score [\%]}]
\addplot[mark=*, mark options={fill=white,magenta},style={magenta,/tikz/mark size=3}] table [x=S, y=F1] {cn_crf_test.data};
\addplot[mark=square*, mark options={fill=black,blue}, style={blue,/tikz/mark size=3}] table [x=S, y=F1] {cn_ge_test.data};
\addplot [green, no markers, dashed] coordinates {(-1,64.40) (16,64.40)};
\end{axis}

\begin{scope}[shift={(9,1)}] 
\draw (0,0) -- 
  plot[mark=*, mark options={fill=white,magenta}, style={magenta,/tikz/mark size=3}] (0.25,0) -- (0.5,0) 
  node[right,font=\huge]{supervised CRF};
\draw[yshift=\baselineskip*2] (0,0) -- 
  plot[mark=square*, mark options={fill=black,blue},style={blue,/tikz/mark size=3}] (0.25,0) -- (0.5,0)
  node[right, font=\huge]{CLiPPER soft};
\end{scope}

\end{tikzpicture}

%% file: de_test.tex

\begin{tikzpicture}
\pgfplotsset{
  tick label style = {font=\Large},
  every axis label = {font=\Large},
  legend style = {font=\huge},
  label style = {font=\huge}
}
\begin{axis}[ width=\linewidth, ytick scale label code/.code={},
    xmin=0, xmax=12,
        xtick=data, xlabel={\# of labeled training sentences [k]}, 
                ylabel={F1 score [\%]}]
\addplot[mark=*, mark options={fill=white,magenta},style={magenta,/tikz/mark size=3}] table [x=S, y=F1] {de_crf_test.data};
\addplot[mark=square*, mark options={fill=black,blue}, style={blue,/tikz/mark size=3}] table [x=S, y=F1] {de_ge_test.data};
\addplot [green, no markers, dashed] coordinates {(-1,59.88) (13,59.88)};
\end{axis}

\begin{scope}[shift={(9,1)}] 
\draw (0,0) -- 
  plot[mark=*, mark options={fill=white,magenta}, style={magenta,/tikz/mark size=3}] (0.25,0) -- (0.5,0) 
  node[right,font=\huge]{supervised CRF};
\draw[yshift=\baselineskip*2] (0,0) -- 
  plot[mark=square*, mark options={fill=black,blue},style={blue,/tikz/mark size=3}] (0.25,0) -- (0.5,0)
  node[right, font=\huge]{CLiPPER soft};
\end{scope}

\end{tikzpicture}

%% file: cn_test_sVSh.tex

\begin{tikzpicture}
\pgfplotsset{
  tick label style = {font=\Large},
  every axis label = {font=\Large},
  legend style = {font=\huge},
  label style = {font=\huge}
}

\begin{axis}[ width=\linewidth, ytick scale label code/.code={},
    xmin=0, xmax=15,
        xtick=data, xlabel={\# of labeled training sentences [k]}, 
                ylabel={F1 score [\%]}]
\addplot[mark=square*, mark options={fill=black,blue}, style={blue,/tikz/mark size=3}] table [x=S, y=F1] {cn_ge_test.data};
\addplot[mark=triangle*, mark options={fill=white,red},style={red,/tikz/mark size=3}] table [x=S, y=F1] {cn_ge_test_hard.data};
\addplot[mark=o, mark options={fill=white,black},style={black,/tikz/mark size=3}] table [x=S, y=F1] {cn_crf_test_hard.data};
\end{axis}

\begin{scope}[shift={(2,3)}] 
\draw (0,0) -- 
  plot[mark=o, mark options={fill=white,black}, style={black,/tikz/mark size=3}] (0.25,0) -- (0.5,0) 
  node[right,font=\huge]{CRF projection};
\draw[yshift=\baselineskip*2] (0,0) -- 
  plot[mark=triangle*, mark options={fill=white,red}, style={green,/tikz/mark size=3}] (0.25,0) -- (0.5,0) 
  node[right,font=\huge]{CLiPPER hard};
\draw[yshift=\baselineskip*4] (0,0) -- 
  plot[mark=square*, mark options={fill=black,blue},style={blue,/tikz/mark size=3}] (0.25,0) -- (0.5,0)
  node[right, font=\huge]{CLiPPER soft};
\end{scope}

\end{tikzpicture}

%% file: de_test_sVSh.tex

\begin{tikzpicture}
\pgfplotsset{
  tick label style = {font=\Large},
  every axis label = {font=\Large},
  legend style = {font=\huge},
  label style = {font=\huge}
}

\begin{axis}[ width=\linewidth, ytick scale label code/.code={},
    xmin=0, xmax=12,
        xtick=data, xlabel={\# of labeled training sentences [k]}, 
                ylabel={F1 score [\%]}]
\addplot[mark=square*, mark options={fill=black,blue}, style={blue,/tikz/mark size=3}] table [x=S, y=F1] {de_ge_test.data};
\addplot[mark=triangle*, mark options={fill=white,red},style={red,/tikz/mark size=3}] table [x=S, y=F1] {de_ge_test_hard.data};
\addplot[mark=o, mark options={fill=white,black},style={black,/tikz/mark size=3}] table [x=S, y=F1] {de_crf_hard_test.data};
\end{axis}

\begin{scope}[shift={(9,1)}] 
\draw (0,0) -- 
  plot[mark=o, mark options={fill=white,black}, style={black,/tikz/mark size=3}] (0.25,0) -- (0.5,0) 
  node[right,font=\huge]{CRF projection};
\draw[yshift=\baselineskip*2] (0,0) -- 
  plot[mark=triangle*, mark options={fill=white,red}, style={green,/tikz/mark size=3}] (0.25,0) -- (0.5,0) 
  node[right,font=\huge]{CLiPPER hard};
\draw[yshift=\baselineskip*4] (0,0) -- 
  plot[mark=square*, mark options={fill=black,blue},style={blue,/tikz/mark size=3}] (0.25,0) -- (0.5,0)
  node[right, font=\huge]{CLiPPER soft};
\end{scope}

\end{tikzpicture}

%% file: CLiPPER.bbl
\begin{thebibliography}{}

\bibitem[\protect\citename{Alshawi \bgroup et al.\egroup
  }2000]{Alshawi:2000:MT}
Hiyan Alshawi, Srinivas Bangalore, and Shona Douglas.
\newblock 2000.
\newblock Head-transducer models for speech translation and their automatic
  acquisition from bilingual data.
\newblock {\em Machine Translation}, 15.

\bibitem[\protect\citename{Ando and Zhang}2005]{Ando:2005:ACL}
Rie~Kubota Ando and Tong Zhang.
\newblock 2005.
\newblock A high-performance semi-supervised learning method for text chunking.
\newblock In {\em Proceedings of ACL}.

\bibitem[\protect\citename{Bellare \bgroup et al.\egroup
  }2009]{Bellare:2009:UAI}
Kedar Bellare, Gregory Druck, and Andrew McCallum.
\newblock 2009.
\newblock Alternating projections for learning with expectation constraints.
\newblock In {\em Proceedings of UAI}.

\bibitem[\protect\citename{Blum and Mitchell}1998]{Blum:1998:COLT}
Avrim Blum and Tom Mitchell.
\newblock 1998.
\newblock Combining labeled and unlabeled data with co-training.
\newblock In {\em Proceedings of COLT}.

\bibitem[\protect\citename{Burkett and Klein}2008]{Burkett:2008:EMNLP}
David Burkett and Dan Klein.
\newblock 2008.
\newblock Two languages are better than one (for syntactic parsing).
\newblock In {\em Proceedings of EMNLP}.

\bibitem[\protect\citename{Burkett \bgroup et al.\egroup
  }2010]{Burkett:2010:CONLL}
David Burkett, Slav Petrov, John Blitzer, and Dan Klein.
\newblock 2010.
\newblock Learning better monolingual models with unannotated bilingual text.
\newblock In {\em Proceedings of CoNLL}.

\bibitem[\protect\citename{Carlson \bgroup et al.\egroup
  }2010]{Carlson:2010:WSDM}
Andrew Carlson, Justin Betteridge, Richard~C. Wang, Estevam R.~Hruschka Jr.,
  and Tom~M. Mitchell.
\newblock 2010.
\newblock Coupled semi-supervised learning for information extraction.
\newblock In {\em Proceedings of WSDM}.

\bibitem[\protect\citename{Chang \bgroup et al.\egroup }2007]{Chang:2007:ACL}
Ming-Wei Chang, Lev Ratinov, and Dan Roth.
\newblock 2007.
\newblock Guiding semi-supervision with constraint-driven learning.
\newblock In {\em Proceedings of ACL}.

\bibitem[\protect\citename{Che \bgroup et al.\egroup }2013]{Che:2013:NAACL}
Wanxiang Che, Mengqiu Wang, and Christopher~D. Manning.
\newblock 2013.
\newblock Named entity recognition with bilingual constraints.
\newblock In {\em Proceedings of NAACL}.

\bibitem[\protect\citename{Collins and Singer}1999]{Collins:1999:EMNLP}
Michael Collins and Yoram Singer.
\newblock 1999.
\newblock Unsupervised models for named entity classification.
\newblock In {\em Proceedings of EMNLP}.

\bibitem[\protect\citename{Das and Petrov}2011]{Das:2011:ACL}
Dipanjan Das and Slav Petrov.
\newblock 2011.
\newblock Unsupervised part-of-speech tagging with bilingual graph-based
  projections.
\newblock In {\em Proceedings of ACL}.

\bibitem[\protect\citename{Druck and McCallum}2010]{Druck:2010:ICML}
Gregory Druck and Andrew McCallum.
\newblock 2010.
\newblock High-performance semi-supervised learning using discriminatively
  constrained generative models.
\newblock In {\em Proceedings of ICML}.

\bibitem[\protect\citename{Druck \bgroup et al.\egroup }2007]{Druck:2007:NIPS}
Gregory Druck, Gideon Mann, and Andrew McCallum.
\newblock 2007.
\newblock Leveraging existing resources using generalized expectation criteria.
\newblock In {\em Proceedings of NIPS Workshop on Learning Problem Design}.

\bibitem[\protect\citename{Druck \bgroup et al.\egroup }2009]{Druck:2009:EMNLP}
Gregory Druck, Burr Settles, and Andrew McCallum.
\newblock 2009.
\newblock Active learning by labeling features.
\newblock In {\em Proceedings of EMNLP}.

\bibitem[\protect\citename{Druck}2011]{Druck:2011:Thesis}
Gregory Druck.
\newblock 2011.
\newblock {\em Generalized Expectation Criteria for Lightly Supervised
  Learning}.
\newblock {Ph.D.} thesis, University of Massachusetts Amherst.

\bibitem[\protect\citename{Faruqui and Pad\'o}2010]{Faruqui:2010:KONVENS}
Manaal Faruqui and Sebastian Pad\'o.
\newblock 2010.
\newblock Training and evaluating a {G}erman named entity recognizer with
  semantic generalization.
\newblock In {\em Proceedings of KONVENS}.

\bibitem[\protect\citename{Finkel \bgroup et al.\egroup
  }2006]{Finkel:2006:EMNLP}
Jenny~Rose Finkel, Christopher~D. Manning, and Andrew~Y. Ng.
\newblock 2006.
\newblock Solving the problem of cascading errors: Approximate bayesian
  inference for linguistic annotation pipelines.
\newblock In {\em Proceedings of EMNLP}.

\bibitem[\protect\citename{Fossum and Abney}2005]{Fossum:2005:IJCNLP}
Victoria Fossum and Steven Abney.
\newblock 2005.
\newblock Automatically inducing a part-of-speech tagger by projecting from
  multiple source languages across aligned corpora.
\newblock In {\em Proceedings of IJCNLP}.

\bibitem[\protect\citename{Ganchev \bgroup et al.\egroup
  }2009]{Ganchev:2009:ICML}
Kuzman Ganchev, Jennifer Gillenwater, and Ben Taskar.
\newblock 2009.
\newblock Dependency grammar induction via bitext projection constraints.
\newblock In {\em Proceedings of ACL}.

\bibitem[\protect\citename{Ganchev \bgroup et al.\egroup
  }2010]{Ganchev:2010:JMLR}
Kuzman Ganchev, Jo\ {a}o Gra\c{c}a, Jennifer Gillenwater, and Ben Taskar.
\newblock 2010.
\newblock Posterior regularization for structured latent variable models.
\newblock {\em JMLR}, 10:2001--2049.

\bibitem[\protect\citename{Goldberg}2010]{Goldberg:2010:Thesis}
Andrew~B. Goldberg.
\newblock 2010.
\newblock {\em New Directions in Semi-supervised Learning}.
\newblock {Ph.D.} thesis, University of Wisconsin-Madison.

\bibitem[\protect\citename{Hovy \bgroup et al.\egroup }2006]{Hovy:2006:NAACL}
Eduard Hovy, Mitchell Marcus, Martha Palmer, Lance Ramshaw, and Ralph
  Weischedel.
\newblock 2006.
\newblock {OntoNotes}: the 90\% solution.
\newblock In {\em Proceedings of NAACL-HLT}.

\bibitem[\protect\citename{Klein}2005]{Klein:2005:Thesis}
Dan Klein.
\newblock 2005.
\newblock {\em The Unsupervised Learning of Natural Language Structure}.
\newblock {Ph.D.} thesis, Stanford University.

\bibitem[\protect\citename{Lafferty \bgroup et al.\egroup
  }2001]{Lafferty:2001:ICML}
John~D. Lafferty, Andrew McCallum, and Fernando C.~N. Pereira.
\newblock 2001.
\newblock Conditional random fields: Probabilistic models for segmenting and
  labeling sequence data.
\newblock In {\em Proceedings of ICML}.

\bibitem[\protect\citename{Li and Eisner}2009]{Li:2009:EMNLP}
Zhifei Li and Jason Eisner.
\newblock 2009.
\newblock First- and second-order expectation semirings with applications to
  minimum-risk training on translation forests.
\newblock In {\em Proceedings of EMNLP}.

\bibitem[\protect\citename{Li \bgroup et al.\egroup }2012]{Li:2012:EMNLP}
Shen Li, Jo\ {a}o Gra\c{c}a, and Ben Taskar.
\newblock 2012.
\newblock Wiki-ly supervised part-of-speech tagging.
\newblock In {\em Proceedings of EMNLP-CoNLL}.

\bibitem[\protect\citename{Liang}2005]{Liang:2005:Thesis}
Percy Liang.
\newblock 2005.
\newblock Semi-supervised learning for natural language.
\newblock Master's thesis, Massachusetts Institute of Technology.

\bibitem[\protect\citename{Mann and McCallum}2010]{Mann:2010:JMLR}
Gideon Mann and Andrew McCallum.
\newblock 2010.
\newblock Generalized expectation criteria for semi-supervised learning with
  weakly labeled data.
\newblock {\em JMLR}, 11:955--984.

\bibitem[\protect\citename{McClosky \bgroup et al.\egroup
  }2006]{McClosky:2006:NAACL}
David McClosky, Eugene Charniak, and Mark Johnson.
\newblock 2006.
\newblock Effective self-training for parsing.
\newblock In {\em Proceedings of NAACL-HLT}.

\bibitem[\protect\citename{Naseem \bgroup et al.\egroup
  }2009]{Naseem:2009:JMLR}
Tahira Naseem, Benjamin Snyder, Jacob Eisenstein, and Regina Barzilay.
\newblock 2009.
\newblock Multilingual part-of-speech tagging: Two unsupervised approaches.
\newblock {\em JAIR}, 36:1076--9757.

\bibitem[\protect\citename{Petrov \bgroup et al.\egroup
  }2010]{Petrov:2010:EMNLP}
Slav Petrov, Pi-Chuan Chang, Michael Ringgaard, and Hiyan Alshawi.
\newblock 2010.
\newblock Uptraining for accurate deterministic question parsing.
\newblock In {\em Proceedings of EMNLP}.

\bibitem[\protect\citename{Samdani \bgroup et al.\egroup
  }2012]{Samdani:2012:NAACL}
Rajhans Samdani, Ming-Wei Chang, and Dan Roth.
\newblock 2012.
\newblock Unified expectation maximization.
\newblock In {\em Proceedings of NAACL}.

\bibitem[\protect\citename{Sang and Meulder}2003]{Sang:2003:CoNLL}
Erik F. Tjong~Kim Sang and Fien~De Meulder.
\newblock 2003.
\newblock Introduction to the {CoNLL-2003} shared task: language-independent
  named entity recognition.
\newblock In {\em Proceedings of CoNLL}.

\bibitem[\protect\citename{Sindhwani \bgroup et al.\egroup
  }2005]{Sindhwani:2005:ICML}
Vikas Sindhwani, Partha Niyogi, and Mikhail Belkin.
\newblock 2005.
\newblock A co-regularization approach to semi-supervised learning with
  multiple views.
\newblock In {\em Proceedings of ICML Workshop on Learning with Multiple Views,
  International Conference on Machine Learning}.

\bibitem[\protect\citename{Smith}2006]{Smith:2006:Thesis}
Noah~A. Smith.
\newblock 2006.
\newblock {\em Novel Estimation Methods for Unsupervised Discovery of Latent
  Structure in Natural Language Text}.
\newblock {Ph.D.} thesis, Johns Hopkins University.

\bibitem[\protect\citename{Snyder \bgroup et al.\egroup }2009]{Snyder:2009:ACL}
Benjamin Snyder, Tahira Naseem, and Regina Barzilay.
\newblock 2009.
\newblock Unsupervised multilingual grammar induction.
\newblock In {\em Proceedings of ACL}.

\bibitem[\protect\citename{Suzuki and Isozaki}2008]{Suzuki:2008:ACL}
Jun Suzuki and Hideki Isozaki.
\newblock 2008.
\newblock Semi-supervised sequential labeling and segmentation using giga-word
  scale unlabeled data.
\newblock In {\em Proceedings of ACL}.

\bibitem[\protect\citename{T\"{a}ckstr\"{o}m \bgroup et al.\egroup
  }2013]{Tackstrom:2013:ACL}
Oscar T\"{a}ckstr\"{o}m, Dipanjan Das, Slav Petrov, Ryan McDonald, and Joakim
  Nivre.
\newblock 2013.
\newblock Token and type constraints for cross-lingual part-of-speech tagging.
\newblock In {\em Proceedings of ACL}.

\bibitem[\protect\citename{Turian \bgroup et al.\egroup }2010]{Turian:2010:ACL}
Joseph Turian, Lev Ratinov, and Yoshua Bengio.
\newblock 2010.
\newblock Word representations: A simple and general method for semi-supervised
  learning.
\newblock In {\em Proceedings of the 48th Annual Meeting of the Association for
  Computational Linguistics (ACL)}.

\bibitem[\protect\citename{Wang \bgroup et al.\egroup }2013]{Wang:2013:AAAI}
Mengqiu Wang, Wanxiang Che, and Christopher~D. Manning.
\newblock 2013.
\newblock Effective bilingual constraints for semi-supervised learning of named
  entity recognizers.
\newblock In {\em Proceedings of AAAI}.

\bibitem[\protect\citename{Xi and Hwa}2005]{Xi:2012:EMNLP}
Chenhai Xi and Rebecca Hwa.
\newblock 2005.
\newblock A backoff model for bootstrapping resources for non-english
  languages.
\newblock In {\em Proceedings of HLT-EMNLP}.

\bibitem[\protect\citename{Yarowsky and Ngai}2001]{Yarowsky:2001:NAACL}
David Yarowsky and Grace Ngai.
\newblock 2001.
\newblock Inducing multilingual {POS} taggers and {NP} bracketers via robust
  projection across aligned corpora.
\newblock In {\em Proceedings of NAACL}.

\bibitem[\protect\citename{Yarowsky}1995]{Yarowsky:1995:ACL}
David Yarowsky.
\newblock 1995.
\newblock Unsupervised word sense disambiguation rivaling supervised methods.
\newblock In {\em Proceedings of ACL}.

\end{thebibliography}
